\documentclass[a4paper, 10pt, conference]{ieeeconf}      
\usepackage{graphicx}
\IEEEoverridecommandlockouts                              

\usepackage{amsmath}
\usepackage{amssymb}
\usepackage{graphics}
\usepackage[hidelinks]{hyperref}
\usepackage{xcolor}
\usepackage{dblfloatfix}
\usepackage {tikz}
\usepackage{mathtools}
\usepackage{svg}
\usepackage{comment}
\usepackage[bottom]{footmisc}
\usepackage{tabularray}
\usepackage{amsmath}
\usepackage{soul}
\usepackage[switch]{lineno}
\usepackage{multirow}
\usepackage{caption}
\usepackage[noend]{algpseudocode}
\usepackage[linesnumbered,ruled,vlined, noend]{algorithm2e}
\usetikzlibrary {positioning}

\title{\LARGE \bf
Relational Q-Functionals: Multi-Agent Learning to Recover from Unforeseen Robot Malfunctions in Continuous Action Domains
}

\author{Yasin Findik$^{1}$, Paul Robinette$^{2}$, Kshitij Jerath$^{3}$, and  Reza Azadeh$^{1}$
\thanks{$^{1}$ PeARL Lab, Richard Miner School of Computer and Information Sciences, University of Massachusetts Lowell, MA, USA {\tt\small yasin\_findik@student.uml.edu, reza@cs.uml.edu}}%
\thanks{$^{2}$ Department of Electrical and Computer Engineering, University of Massachusetts Lowell, MA, USA {\tt\small paul\_robinette@uml.edu}}%
\thanks{$^{3}$ Department of Mechanical Engineering, University of Massachusetts Lowell, MA, USA {\tt\small kshitij\_jerath@uml.edu}}
}

\begin{document}

\maketitle
\thispagestyle{empty}
\pagestyle{empty}

\begin{abstract}
Cooperative multi-agent learning methods are essential in developing effective cooperation strategies in multi-agent domains. In robotics, these methods extend beyond multi-robot scenarios to single-robot systems, where they enable coordination among different robot modules (e.g., robot legs or joints). However, current methods often struggle to quickly adapt to unforeseen failures, such as a malfunctioning robot leg, especially after the algorithm has converged to a strategy. To overcome this, we introduce the Relational Q-Functionals (RQF) framework. RQF leverages a relational network, representing agents' relationships, to enhance adaptability, providing resilience against malfunction(s). Our algorithm also efficiently handles continuous state-action domains, making it adept for robotic learning tasks. Our empirical results show that RQF enables agents to use these relationships effectively to facilitate cooperation and recover from an unexpected malfunction in single-robot systems with multiple interacting modules. Thus, our approach offers promising applications in multi-agent systems, particularly in scenarios with unforeseen malfunctions.

\end{abstract}

\section{Introduction}

Multi-agent learning methods have gained significant attention for their ability to handle the complexities of robotic systems~\cite{dorri2018multi}. These methods, which are crucial for tasks that require intricate coordination, are applicable not only in scenarios involving multiple robots but also in single-robot systems. In single-robot applications, multi-agent learning algorithms can be beneficial by learning coordination among different modules of the system. For instance, each joint of a robot manipulator can be considered as an individual agent within a team of agents. The multi-agent learning algorithms enable the agents (i.e., joints) to working collaboratively with other agents (i.e., joints) to discover an optimal policy for executing a given task. The importance of such collaborative behaviors is particularly highlighted during unexpected robot malfunctions, like joint failures. In these situations, agents must collaborate and adapt their strategies to overcome these challenges.

Recent advancements in deep reinforcement learning have demonstrated considerable potential in tackling the complexities of cooperation in multi-agent settings~\cite{huttenrauch2019deep, levine2016end, mnih2015human}. Notably, a key paradigm in multi-agent reinforcement learning, known as centralized training with decentralized execution (CTDE), has emerged as a significant approach for handling cooperative tasks~\cite{oliehoek2008optimal}. This paradigm adeptly navigates various cooperative challenges, such as the curse of dimensionality as highlighted in~\cite{shoham2007if}, non-stationarity as discussed in~\cite{busoniu2008comprehensive}, and issues related to global exploration~\cite{matignon2012independent}. Yet, adapting slowly to unexpected robot failures and malfunctions is a challenging problem~\cite{findik2023collaborative}. The limitations in the adaptation capabilities of the CTDE-based methods can be attributed primarily to two factors: (i) they are not specifically built to handle such robot failures, (ii) the inherent design of these methods lacks components that enhance cooperation among agents towards a certain strategy.

In this study, we propose a novel multi-agent learning algorithm that integrates a relational network, underscoring the relative importance among agents, into the CTDE paradigm. The main goal of our algorithm is to expedite behavior adaptation in unexpected robot failure scenarios. One of the main factors that makes our algorithm suitable for robot learning tasks is that it can handle continuous state-action domains efficiently. We evaluate our algorithm in four experiments within a simulated multi-agent setting for handling malfunctions in the ant robot environment~\cite{gymnasium_robotics2023github}. Our results show that our approach not only fosters more effective cooperation among agents (i.e., robot legs) but also accelerates adaptation to unforeseen situations by leveraging the power of relational networks.

\section{Related Work}

The field of Multi-Agent Reinforcement Learning (MARL), particularly in cooperative domains, has seen significant advancement in recent years. This progress has largely been driven by diverse approaches aimed at fostering effective collaboration among agents to achieve shared goals. Centralized learning, where all agents share a single controller to learn a joint policy or value function, is a notable approach in this context~\cite{claus1998dynamics}. However, despite its potential benefits, this approach faces computational challenges and intractability issues, particularly as the number of agents and the complexity of their interactions grow. An alternative to centralized learning is the decentralized learning approach, wherein each agent independently develops its own policy. This approach, exemplified by techniques like Independent Q-Learning (IQL)~\cite{tan1993multi}, enables cooperative behavior to emerge through the application of these policies within the environment. Extensions to IQL have incorporated function approximation to manage the complexity of high-dimensional spaces~\cite{tampuu2017multiagent}. 
However, decentralized learning faces challenges of non-stationarity, a consequence of the evolving actions of other agents, and a potential breach of the Markov property, undermining the reliability of convergence in these algorithms~\cite{hernandez2017survey}.

Addressing the limitations inherent in both fully centralized and decentralized learning, the Centralized Training with Decentralized Execution (CTDE)~\cite{oliehoek2008optimal} paradigm has emerged as a novel approach in cooperative MARL scenarios. CTDE combines the benefits of centralized training, mitigating non-stationarity issues, with decentralized execution to ensure scalability. This paradigm has been employed through two primary methodologies: (i) policy-based methods, like Multi-Agent Deep Deterministic Policy Gradient (MADDPG)~\cite{lowe2017multi} and Multi-Agent Proximal Policy Optimization (MAPPO)~\cite{yu2022surprising}, which involve a centralized critic aware of all agents' global observations, and (ii) value-based methods, such as Value Decomposition Networks (VDN)~\cite{sunehag2017value}, QMIX~\cite{rashid2020monotonic}, and QTRAN~\cite{son2019qtran}, that use a centralized function to derive joint Q-values from individual agent action-values. These methods have demonstrated their efficacy in resolving multi-agent coordination challenges across various settings.

Despite the progress in cooperative MARL, existing strategies primarily target optimal solutions under normal conditions and may lack prompt adaptability during unforeseen failures. To address agent malfunctions, one potential approach might be predicting such events and their timing, potentially using concepts like LOLA~\cite{foerster2017learning} to enhance performance. However, in cases where malfunctions are unpredictable, the challenge shifts to improving agents' capacity for rapid adaptation.

In this paper, we introduce a novel framework designed to facilitate collaborative adaptation in the face of unexpected agent malfunctions. Our methodology is centered around the use of inter-agent relational networks~\cite{findik2023impact,findik2023influence,haeri2022reward}, capturing the level of importance that agents assign to each other. By leveraging these networks, agents can swiftly adjust their behaviors to compensate for the sudden failure of teammates. We investigate this idea through the lens of value-based concept within continuous action domains. It is important to note, however, that our proposed framework, which utilizes relationships to navigate unforeseen malfunctions, has potential applicability across various policy-based methods.

\section{Background}

\subsection{Markov Decision Process}
We characterized Markov Decision Process as a tuple $\langle  \mathcal{S}, \mathcal{A}, \mathcal{R}, \mathcal{T}, N, \gamma\rangle$ where $s \in \mathcal{S}$ indicates the true state of the environment, $\mathcal{A}\coloneqq \{a_1, a_2, \dots, a_n \}$ and $\mathcal{R}\coloneqq \{r_1, r_2, \dots, r_n \}$ are the set of agents' action and reward,  respectively. $\mathcal{T} (s, \mathcal{A}, s') \colon \mathcal{S} \times \mathcal{A}\times \mathcal{S} \mapsto [0,1]$ is the transition function, where $s'$ is the next state. $N$ denotes the number of agents and $\gamma\in[0,1)$ is the discount factor.

\subsection{Q-Learning}

Tabular Q-Learning, initially introduced in~\cite{watkins1992q}, utilizes an action-value $Q^{\pi}$ for a policy $\pi$, defined as 
$Q^{\pi}(s, a)=\mathbb{E}[G| s=s_t, a=a_t]$, where $G$ and $t$ represent the return and time-steps, respectively. And, it can be reformulated recursively as follows:
$$
Q^{\pi}(s, a)=\mathbb{E}_{s'\sim{\pi(s,a)}}[R(s,a) + \gamma \mathbb{E}_{a'\sim\pi}[Q^{\pi}(s', a')]],
$$
\noindent where $R$ denotes the reward function.

On the other hand, Deep Q-Network (DQN)~\cite{mnih2015human} is a variant Q-Learning that utilizes a neural network to approximate the action-value function, represented as $\hat{Q}(s, a; w)$ 
where $w$ is the parameters of the neural network, by minimizing the loss:
\begin{align}
\label{loss_DQN}
\resizebox{.91\linewidth}{!}{$
            \displaystyle
            L(w_{PN})=\mathbb{E}_{s,a,r,s'}[(r + \gamma \max_{a'}(\hat{Q}(s', a'; w_{\textrm{TN}}))- \hat{Q}(s, a; w_{\textrm{PN}}))^2],
        $}
\end{align}
\noindent 
where $w_{\textrm{PN}}$ and $w_{\textrm{TN}}$ are the parameters of the prediction and target networks, respectively. DQN enhances the stability of the learning process by periodically updating the target network parameters $w_{\textrm{TN}}$ with those of the prediction network $w_{\textrm{PN}}$. Additionally, DQN incorporates an experience replay buffer, which stores tuples in the form of $\langle s, a, r, s'\rangle$, to further strengthen the model's stability.

Most value-based methods, including Q-Learning and DQN, have demonstrated their efficacy in deriving robust policies for agents operating within environments characterized by discrete action spaces, where the number of possible actions is finite. In contrast, these methods frequently encounter challenges when applied to continuous environments. In such settings, the action space is represented by continuous vectors, leading to an infinitely large set of potential actions. This characteristic considerably diminishes the effectiveness of value-based methods in formulating viable policies for such settings~\cite{lim2018actor}. This limitation is particularly important considering that most real-world robotic applications predominantly operate within continuous action domains. 

\subsection{Q-functionals}

Q-functionals~\cite{lobel2023q}, which our proposed method builds upon, renders the value-based concept highly effective in continuous action domains,  outperforming policy-based methods. The essence of Q-functionals lies in the innovative representation of Q-functions, enabling the concurrent evaluation of multiple action values for a given state. Traditional DQN implementations establish a direct mapping from state-action pairs to $\mathcal{R}$, as defined in:
\begin{align}
\label{tqa}
\hat{Q}(s,a): (\mathcal{S} \times \mathcal{A}) \mapsto \mathcal{R}.
\end{align}
\noindent
On the other hand, Q-functionals propose a novel mapping mechanism, in which the states are initially mapped to a set of basis functions. These basis functions are then utilized to map actions to the set of $\mathcal{R}$, as follows:
\begin{align}
\label{fqa}
\hat{Q}^{\textrm{F}}(s,a): \mathcal{S} \mapsto (\mathcal{A} \mapsto \mathcal{R}).
\end{align}
In conventional architectures, as outlined in~\eqref{tqa}, the computational demand for evaluating action values for two actions in the same state is approximately double that required for a single action. Q-functionals, however, innovate by transforming each state into a parameter set that defines a function over the action space. Here, each state is encoded as a function, formulated through the learned coefficients of basis functions within the action space. These state-representative functions facilitate rapid evaluation of multiple actions via matrix operations involving the actions and the learned coefficients. Consequently, Q-functionals offer an efficient solution for managing environments with continuous action spaces in single-agent scenarios, maintaining high sample efficiency.

\subsection{Independent Q-Functionals}

Among value-based methods for reinforcement learning, Q-functionals play a pivotal role, particularly in handling continuous action spaces. However, Q-functionals' design is primarily oriented towards single-agent tasks. An advancement in this field is seen with the introduction of Independent Q-Functionals (IQF), an extension of Q-functionals as proposed in~\cite{findik2024mixed}. IQF facilitates multi-agent scenarios within continuous environments, enabling the utilization of Q-functionals in a fully decentralized manner. This approach allows for the emergence of cooperative behavior among agents through the application of independently learned policies.

Essentially, IQF, similar to the principles of IQL~\cite{tan1993multi}, seeks to independently learn an optimal functional, denoted as ${\hat{Q}}^{\textrm{F}*}_i$ in~\eqref{fqa}, for each agent. This independent learning mechanism, wherein agents learn and deploy their policies without exchanging information, provides IQF with a robust scalability feature. Nonetheless, akin to IQL, IQF might be vulnerable to non-stationarity issues. These issues are a consequence of the dynamic nature of the actions of other agents within the environment, as perceived by an individual agent. The extent of these non-stationary challenges varies based on the complexity of the specific task and the cooperation scenarios involved. In our experimental setup, we tackle a relatively uncomplicated problem that is solvable through IQF. Our primary objective is to elucidate the effects of agent malfunctions and the emergent behaviors of agents in response to such events.

\renewcommand{\algorithmicrequire}{\textbf{Input:}}
\renewcommand{\algorithmicensure}{\textbf{Output:}}
\SetAlCapNameFnt{\scriptsize}
\SetAlCapFnt{\scriptsize}

\begin{algorithm}[t]
    \scriptsize
    \caption{Relational Q-Functionals}       
    \label{alg:RQF}
    \SetAlgoLined
    \SetKwInOut{Input}{input}
    \SetKwInOut{Output}{output}
    \DontPrintSemicolon
    
    \Input{prediction network, $\hat{Q}^{\textrm{F}}_{\textrm{prediction}}$; target network, $\hat{Q}^{\textrm{F}}_{\textrm{target}}$; relational graph, $G$; batch size, $b$; update frequency, \textit{step}$_\textrm{update}$; action range, $[\textit{a}_\textrm{min}, \textit{a}_\textrm{max}]$; order of the basis function, $o$;}
    \vspace{0.05cm}
    \ForEach{episode}{
        \vspace{0.05cm}
        Initialize s\Comment{$s \in \mathbb{R}^N$}\;
        \vspace{0.05cm}
        \ForEach{step \textup{\textbf{of}} episode}{

            $a_\textrm{rand}$ $\leftarrow$ $\mathcal{U}$([$\textit{a}_\textrm{min}$, $\textit{a}_\textrm{max}$])\Comment{random action selection}\;
            \vspace{0.05cm}

            $C_s$ $\leftarrow \hat{Q}^{\textrm{F}}_{\textrm{prediction}}$(s)\Comment{calculate state function's coefficients}\;

            \vspace{0.05cm}

            $V_s$ $\leftarrow$ $\Phi$($o$, $a_\textrm{rand}$)\Comment{state function's representation}\;

            \vspace{0.05cm}

            $Q$ $\leftarrow$ $C_s V_s$\Comment{calculate action values using matrix multiplication}\;
            
            \vspace{0.05cm}
            $a_\textrm{best}$ $\leftarrow$ $\arg \max$($Q$)\Comment{best action for each agent}\;

            \vspace{0.05cm}
            $a$ $\leftarrow$  $a_{\textrm{best}} + \epsilon$ \Comment{$\epsilon \sim \mathcal{N}(0, 0.1)$}\;
            
            \vspace{0.05cm}
            Take actions, $a$, observe $r$, $s'$\;
            
            \vspace{0.05cm}
            Store $s$, $a$, $r$, $s'$ in memory\;
            
            \vspace{0.05cm}
            
            $s$ $\leftarrow$  $s'$\;
            
            \vspace{0.05cm}
            \textbf{if} mod(\textit{step}, \textit{step}$_\textrm{update}$) $=0$ \textbf{then} \Comment{network update process}

                        \vspace{0.05cm}
                        \hspace{\algorithmicindent}$S$, $A$, $R$, $S'$ $\leftarrow$ sample chunk, size of $b$, from memory\;
                        
                        \vspace{0.02cm}
                        \hspace{\algorithmicindent}$C_s$ $\leftarrow \hat{Q}^{\textrm{F}}_{\textrm{prediction}}$($S$)\;
                        
                        \vspace{0.05cm}
                        \hspace{\algorithmicindent}$V_s$ $\leftarrow$ $\Phi$($o$, $A$)\;
                        
                        \vspace{0.05cm}
                        \hspace{\algorithmicindent}$Q^\textrm{prediction}\leftarrow$ $C_s V_s$\;
                        
                        \vspace{0.05cm}
                        \hspace{\algorithmicindent}$Q^\textrm{prediction}_{\textrm{team}} \leftarrow$ use~\eqref{action_value_team} with $G$ and $Q^\textrm{prediction}$\;
                        
                        \vspace{0.05cm}
                        \hspace{\algorithmicindent}$a_\textrm{rand}$ $\leftarrow$ $\mathcal{U}$([$\textit{a}_\textrm{min}$, $\textit{a}_\textrm{max}$])\;
                        
                        \vspace{0.05cm}
                        \hspace{\algorithmicindent}$C_s$ $\leftarrow \hat{Q}^{\textrm{F}}_{\textrm{target}}$($S'$)\;
                        
                        \vspace{0.05cm}
                        \hspace{\algorithmicindent}$V_s$ $\leftarrow$ $\Phi$($o$, $a_\textrm{rand}$)\;
                        
                        \vspace{0.05cm}
                        \hspace{\algorithmicindent}$Q^\textrm{target}$ $\leftarrow$ $C_s V_s$\;
                        
                        \vspace{0.05cm}
                        \hspace{\algorithmicindent}$Q^\textrm{target}_\textrm{best} \leftarrow$ $\arg \max$($Q^\textrm{target}$)\;
                        
                        \vspace{0.1cm}
                        \hspace{\algorithmicindent}$Q^\textrm{target}_{\textrm{team}} \leftarrow$ use~\eqref{action_value_team} with $G$ and $Q^\textrm{target}_\textrm{best}$\;
                        
                        \vspace{0.1cm}
                        \hspace{\algorithmicindent}$L \leftarrow$ use~\eqref{loss_DQN} with $R$, $Q^{\textrm{target}}_{\textrm{team}}$, $Q^{\textrm{prediction}}_{\textrm{team}}$\;
                        
                        \vspace{0.05cm}
                        \hspace{\algorithmicindent}Backpropagate $L$ to the parameters of $\hat{Q}^{\textrm{prediction}}$\;
                        
                        \vspace{0.05cm}
                        \hspace{\algorithmicindent}Update weights of $\hat{Q}^{\textrm{F}}_{\textrm{target}}$ with those of $\hat{Q}^{\textrm{F}}_{\textrm{prediction}}$ using ~\eqref{soft_update}\;
                        \vspace{0.05cm}

        }        
    }

\end{algorithm}

\section{Proposed Method}

Various techniques, such as value-based or policy-based, are employed with the objective of maximizing collective rewards in cooperative MARL. These methods facilitate convergence towards one of several potential optimal solutions, possibly including the global optimum. The process of convergence is notably influenced by the stochastic exploration behaviors of individual agents, especially in scenarios where multiple cooperative strategies yield equivalent maximum team reward. On the other hand, the configuration of team plays a pivotal role in determining which cooperative strategies prove effective. This aspect becomes particularly important in real-world applications as they introduce unique challenges. For instance, robots that have adapted to a specific cooperative strategy may suffer unforeseen mechanical issues such as battery or rotational failures. These situations demand a learning approach that is both adaptable and deeply informed by the team's structural dynamics, ensuring the readiness to shift strategies as necessary.

In order to address these challenges effectively, it is imperative to develop a mechanism that takes into consideration the dynamics among agents, and provides guidance for the team to adopt specific cooperation strategies. Such a mechanism holds significant promise for enhancing team performance and expediting adaptation processes, as it can steer agents towards either assisting a malfunctioning agent in resolving its task or completing the task on its behalf. Unfortunately, existing cooperative MARL approaches do not incorporate such a mechanism, resulting in increased complexity and time requirements when adapting to unforeseen malfunctions. To address this critical issue, we introduce a novel algorithm known as Relational Q-Functionals (RQF). RQF enables agents to comprehend inter-agent relationships and make informed decisions about cooperative strategies, allowing them to collectively navigate the challenges associated with adapting to new environmental conditions. In our research, we investigate the impact of inter-agent relationships by utilizing Q-functionals as the learning algorithm for agents, allowing them to apply value-based concepts to solve tasks in a continuous action space.

\begin{figure*}[t]
\centering
\includegraphics[width=0.8\textwidth]{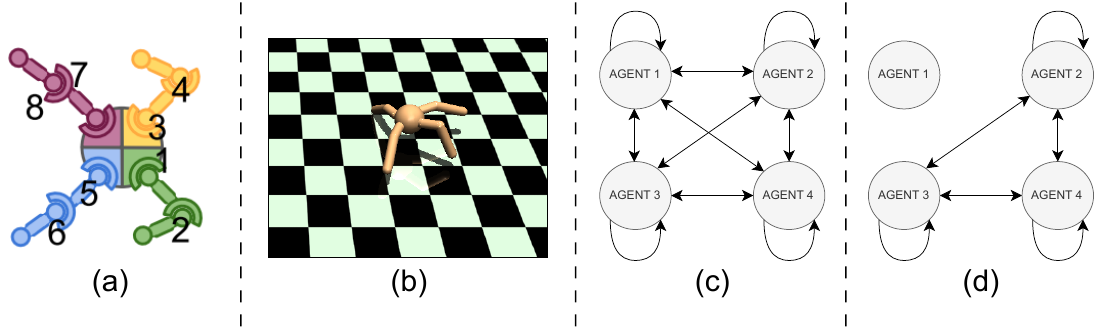}
  \caption{\small{(a) Representation of an ant featuring four agents, each distinguished by a different color. (b) The MaMuJoCo-Ant simulation environment. (c-d) Relational networks used in RQF.}} 
  \label{fig:relations}
\end{figure*}

Relational Q-Functionals employ a relational network structure represented as a directed graph, $\mathcal{G}$, to elucidate the inter-agent relationships. To encourage cooperation among agents, and steer team's cooperation strategy, the team's action value, $Q_{\textrm{team}}$, is calculated according to this relational graph. The relational graph, $\mathcal{G}$, is defined as $\mathcal{G}=(\mathcal{V}, \mathcal{E}, \mathcal{W})$, where each agent, $i \in \{1, 2, ..., N\}$, is represented as a vertex $v_i \in \mathcal{V}$, $\mathcal{E}$ is the set of directed edges labelled as $e_{ij} \in \mathcal{E}$ originated from $v_i$ to $v_j$, and $\mathcal{W}$ encompasses the weight of these edges as $w_{ij} \in \mathcal{W}$ within the range of $[0, 1]$. It is important to note that both the direction of an edge, $e_{ij}$, and its weight, $w_{ij}$, show the significance or placed interest that agent $i$ attributes to the action-value of agent $j$. 

The calculation of the team's action value, $Q_{\textrm{team}}$, is determined by the following expression:
\begin{align}
\label{action_value_team}
Q_{\textrm{team}} = \sum_{i\in\mathcal{V}}^{} \sum_{j\in\mathcal{E}_i}^{} w_{ij}Q_j,
\end{align}
\noindent where the action value $Q_j$ for each individual agent is computed using Q-functionals, wherein a deep neural network is responsible for calculating the coefficients of each state's basis function and performing matrix multiplication with action representations. 

It is important to note that, similar to many other cooperative MARL approaches, our proposed framework also utilizes a replay memory and target network structure in our framework, to increase the stability of the prediction network. The weights of the target network are updated at each time-step using \textit{soft} update mechanism with the prediction network's  weights, employing a small factor ($\tau \ll 1$). As discussed in~\cite{lillicrap2015continuous}, this method is more effective in continuous action domains compared to the periodic updates, which are commonly used in discrete environment. Soft updates can be formulated as:
\begin{equation}
\label{soft_update}
    w_{\textrm{TN}} = \tau w_{\textrm{PN}} + (1-\tau)w_{\textrm{TN}},
\end{equation}
\noindent where $w_{\textrm{TN}}$ and $ w_{\textrm{PN}}$ are the parameters of target network and prediction network, respectively. Algorithm~\ref{alg:RQF} represents the pseudo-code for our proposed method.

\section{Experiments}
\subsection{Environment}

To assess the impact of the proposed method on agent behavior and adaptability in response to unexpected robot failures, we conducted experiments within a multi-agent environment, specifically MaMuJoCo-Ant, introduced in~\cite{peng2021facmac}. The experiments are designed to evaluate the agents' ability to effectively synchronize actions and adapt to unforeseen failures that their teammates might encounter.

The MaMuJoCo-Ant, as illustrated in Fig.~\ref{fig:relations}(b), is a robotic simulation environment that resembles a 3D ant. It features a freely rotating torso connected to four legs, each comprising two joints. The primary objective of this environment is to achieve coordinated movement in the forward direction (towards positive direction in $x$). This is accomplished by strategically applying torques to the eight hinges, enabling control over the ant's movements. Specifically, we utilized a variant that assigns an individual agent to each leg, as shown in Fig~\ref{fig:relations}(a). In other words, the ant is controlled by 4 different agents, each responsible for a leg with two joints. Each agent, in this setting, has two action values, ranging from $[-1, +1]$. During an episode, agents receive a team reward at each step, calculated as $r_\textrm{team} = r_\textrm{stable} + r_\textrm{forward} - r_\textrm{ctrl\_cost}$, where $r_\textrm{stable}$ is the rewards given at every time-step if the ant maintains a stable posture (not upside down), $r_\textrm{forward}$ is the reward of moving forward, measured as $\frac{\Delta x}{dt}$ (with $dt$ being the time between actions and $\Delta x$ is the change in $x$ direction), and $r_\textrm{ctrl\_cost}$ is the penalty for taking excessively large actions. In addition to this reward structure, we penalize the agents with $-100$ reward, if the ant becomes upside down. For our experiments, we set $r_\textrm{stable}$ as $+0.01$ and maximum number of steps per episode as $100$.

\subsection{Models and Hyperparameters}

In our experiments, we employed a Multi-Layer Perceptron (MLP) architecture for each agent's Q-functional representation. These MLPs comprised three hidden layers, each featuring $256$ neurons and utilizing the TanH activation function. The input of these networks is corresponding agent's state, and the output consists of coefficients of a basis function that represents the given state. The size of output can be calculated via the combinatorial formula ${o + d \choose d}$, where $d$ is the action size and $o$ represents the order of a basis function. Specifically, we employed \textit{polynomial} basis functions of order $o=2$ for both RQF and IQF in our experiments.

The training of each agent's coefficient prediction network involves updating its weights every 10 steps, using batches of size $b=512$ randomly sampled from a replay memory with a capacity of $500$k time-steps. We used the \textit{Adam} optimizer with a learning rate of 0.0001, and the squared Temporal Difference (TD) error as the loss function. Additionally, for \textit{soft} target network updates, we set $\tau$ as $0.01$.

For the agents' exploration strategy, we utilized a combination of $\varepsilon$-greedy and Gaussian exploration methods. The $\varepsilon$ value linearly decreased over time, and Gaussian noise was added to the agents' actions to enhance exploration. Lastly, we set the discount factor ($\gamma$) at 0.99, allowing the reinforcement learning process to account for future rewards.

\subsection{Results \& Discussion}

\begin{figure}[b]
\centering
\includegraphics[width=\linewidth]{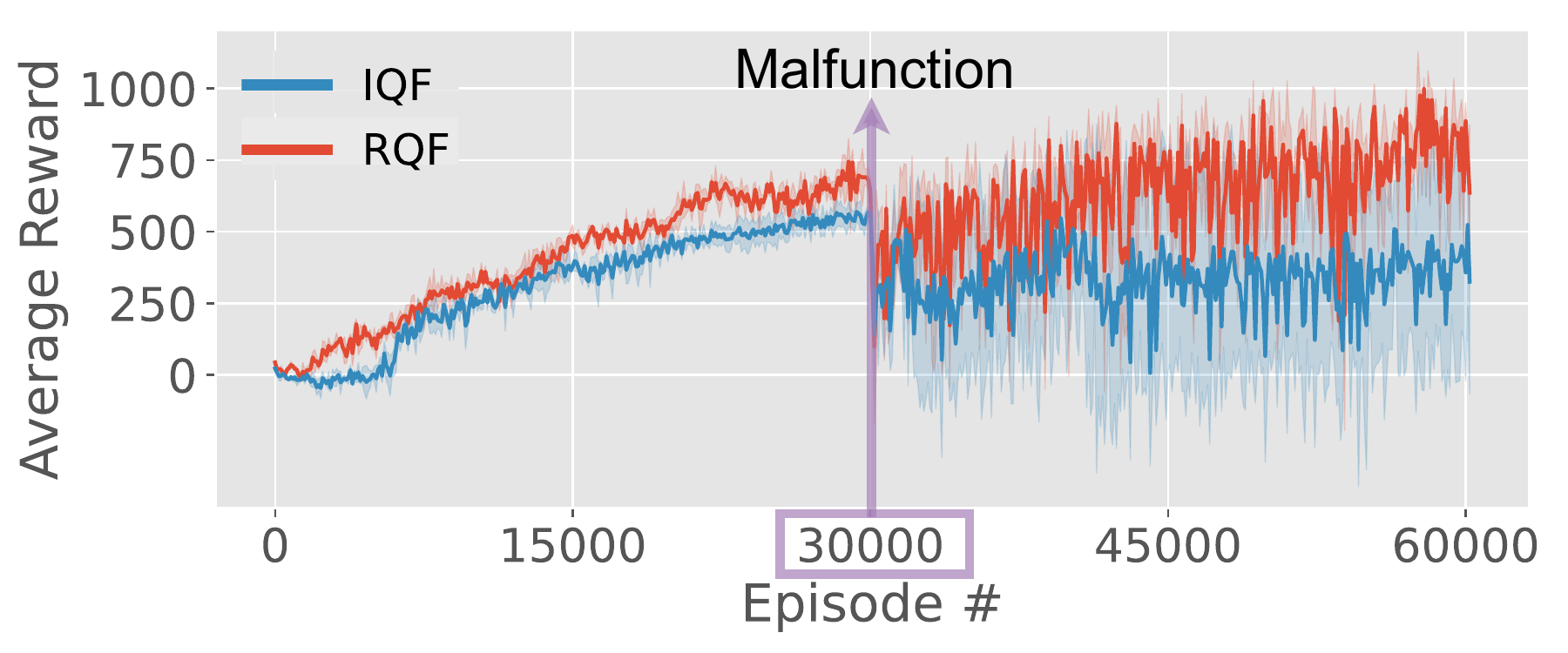}
  \caption{\small{Average team rewards before and after malfunction occurred at the $30000$th episode.}} 
  \label{fig:results}
\end{figure}

\begin{figure}[t]
\centering
\includegraphics[width=\linewidth]{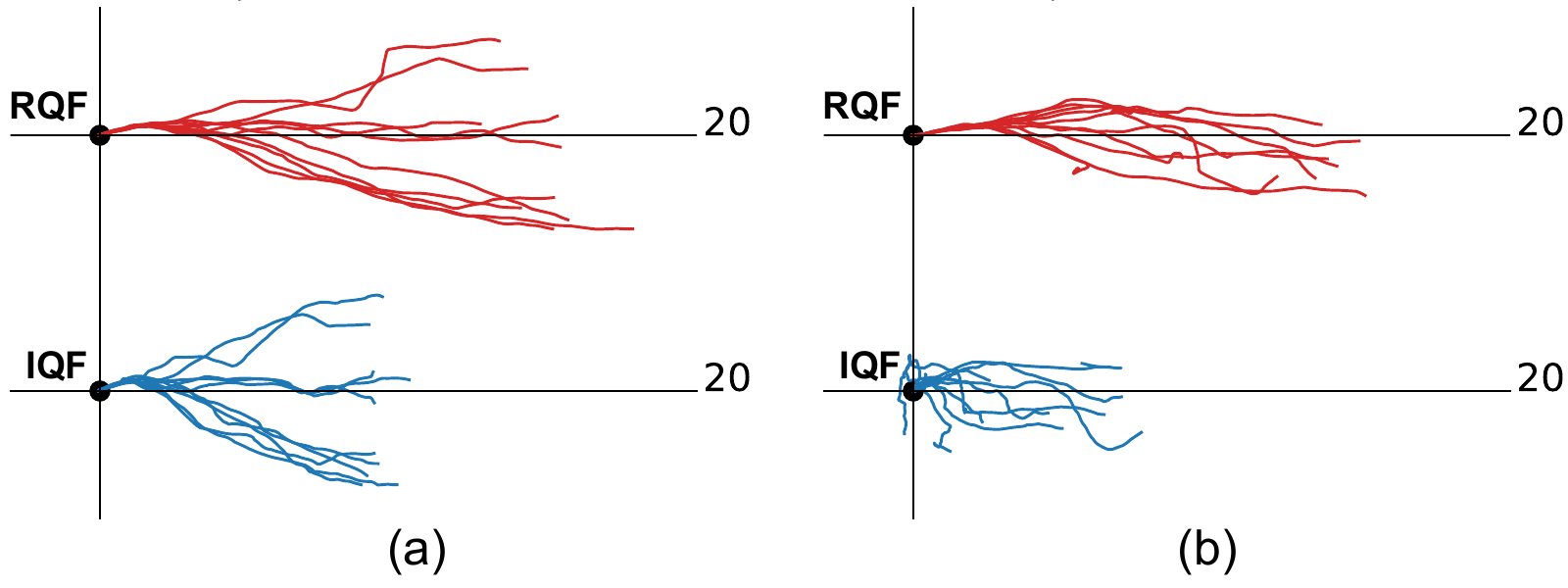}
  \caption{\small{Robot trajectories in x-y plane: (a) before and (b) after malfunction, upon completing 30k and 60k training episodes, respectively. It can be seen that the robot can cover more distance using RQF even after malfunction.}} 
  \label{fig:trajectories}
\end{figure}

The experimental results, depicted in Fig.~\ref{fig:results}, illustrate the average team reward across three runs, with the shaded areas indicating 95\% confidence interval, and the solid lines denoting the average test rewards. These test rewards are assessed using a greedy strategy, pausing training every $100$ episodes to conduct $100$ test episodes, thereby evaluating the collective rewards of the team. Notably, at the thirty thousandth episode of each run, we simulate a malfunction that prevents the movement of one of the ant's legs. This malfunction is not pre-programmed into the agents' learning process but is introduced as an unanticipated challenge. Fig~\ref{fig:trajectories} shows the trajectories of the controlled ant, illustrating the positions of the ant's center in the $x$ and $y$ directions at each time step. The starting locations of the ant for each case are marked by a black dot. These trajectories are obtained by testing ten episodes with the latest trained models of IQF and RQF, both before and after the malfunction occurred. Table~\ref{experiment_results} presents the average team reward and the percentage of ant remaining stable (not being upside down), based on the outcomes from $1000$ test episodes for both before and after malfunction upon the completion of $30$k and $60$k training episodes, respectively.

In the initial phase, wherein none of the agents (i.e., legs) face a malfunction, RQF integrates the relational network depicted in Fig.~\ref{fig:relations}(c). As illustrated in Fig.~\ref{fig:results}, RQF slightly outperforms IQF in terms of team rewards. However, the trajectories presented in Fig.~\ref{fig:trajectories}(a) reveal a noteworthy distinction: agents trained using RQF demonstrate behaviors that lead to more extensive navigation in the $x$ direction. This finding underscores the efficacy of employing a relational network in fostering cooperation among agents under normal conditions (without any malfunction), thus more efficiently directing the ant towards the target direction.

In the event of a malfunction, our approach presupposes the existence of a detection mechanism capable of identifying both the timing of the malfunction (potentially through monitoring the collective reward of the team) and the specific malfunctioning agent. This identification is achieved by analyzing the agents' prior movements and recognizing deviations from established behavioral patterns during training. Upon the occurrence of a malfunction, we modify the relational network employed in training RQF, as depicted in Fig.~\ref{fig:relations}(d). This modification involves diminishing the significance of the malfunctioning agent, considering that its compromised functionality renders its action-value insignificant.

As illustrated in Fig.~\ref{fig:results}, while IQF fails to adapt to the malfunction, RQF, equipped with the adjusted inter-agent relationships, demonstrates an ability to recover from such unforeseen failures. The trajectories depicted in Fig.~\ref{fig:trajectories}(b) demonstrate that  while RQF consistently maintains its orientation towards the target direction, IQF fails to sustain the level of performance it exhibited prior to the malfunction. Crucially, this study underscores that agents trained with the RQF not only show cooperative behaviors, leading to more effective forward movement, but also demonstrate an improved ability to adapt their policies to recover from unexpected malfunctions, leveraging the dynamics of inter-agent relationships.

\begin{table}[b]
\centering
\caption{
\small{Average metrics with 95\% confidence intervals for three runs upon training completion.
}}
\resizebox{\linewidth}{!}{
\begin{tblr}{
  cells = {c},
  cell{1}{2} = {c=2}{},
  cell{1}{4} = {c=2}{},
  vline{3} = {1}{},
  vline{4} = {1-9}{},
  hline{1,10} = {-}{},
}
            & Before Malfunction &              & After Malfunction &              \\
            & Team Reward    & Remaining Stable & Team Reward    & Remaining Stable  \\
IQF   & 566.43 ± 15.64                 & \textbf{0.98 ± 0.00} & 380.76 ± 217.52               & 0.87 ± 0.08  \\
RQF (ours)   & \textbf{663.74 ± 48.24}                 & 0.87 ± 0.05  & \textbf{732.63 ± 51.19}                & \textbf{0.88 ± 0.05} \\\hline       
\end{tblr}
}
\label{experiment_results}
\end{table}

\section{Conclusions and Future Work}

\addtolength{\textheight}{-4.5cm}
In this study, we introduced a novel framework designed to integrate inter-agent relationships into the learning process of agents. Our approach has demonstrated efficacy in enabling cooperative team behaviors in response to unexpected robotic malfunctions. Our experimental results have shown that our method significantly accelerates adapting to unforeseen failures. The ideas presented in this research can potentially be extended to managing recovery in more complex scenarios. This is due to the fact that the proposed framework is more sample efficient (compared to its policy-based counterparts) and can be applied to tasks in continuous action-space domains. Future work will focus on extending experiments to more intricate scenarios involving multiple agents experiencing various unforeseen malfunctions (e.g., noisy or adversarial leg/agent). Additionally, we plan to conduct a comparative analysis against other leading methodologies to further evaluate the robustness and efficiency of our algorithm in such settings.

\section*{Acknowledgement}
This work is supported in part by NSF (IIS-2112633) and
the Army Research Lab (W911NF20-2-0089).

\bibliographystyle{IEEEtran}
\bibliography{IEEEabrv, refs}

\end{document}